\documentclass{article}

\PassOptionsToPackage{numbers, compress}{natbib}
\usepackage[final]{nips_2017}

\usepackage[utf8]{inputenc} 
\usepackage[T1]{fontenc}    
\usepackage{hyperref}       
\usepackage{url}            
\usepackage{booktabs}       
\usepackage{amsfonts}       
\usepackage{nicefrac}       
\usepackage{microtype}      

\usepackage{algorithm}
\usepackage[noend]{algpseudocode}
\usepackage{bbm}
\usepackage{comment}

\usepackage{amsmath,amssymb,amsfonts,graphicx,amsthm,nicefrac,mathtools,bm}

\usepackage{enumitem}
\setlist[itemize,1]{leftmargin=2.5em}

\def\R{\mathbb{R}}

\def\independenT#1#2{\mathrel{\rlap{$#1#2$}\mkern2mu{#1#2}}}
\newcommand\independent{\protect\mathpalette{\protect\independenT}{\perp}}

\newcommand{\T}{\mathcal{T}}
\newcommand{\E}{\mathbb{E}}
\newcommand{\X}{\mathcal{X}}
\newcommand{\Y}{\mathcal{Y}}
\newcommand{\bY}{\mathbf{Y}}
\newcommand{\by}{\mathbf{y}}

\usepackage{tikz}
\usetikzlibrary{arrows}
\usetikzlibrary{shapes}

\newtheorem{theorem}{Theorem}

\newtheorem{lemma}[theorem]{Lemma}

\newtheorem{corollary}[theorem]{Corollary}

\makeatletter
\newcommand{\newreptheorem}[2]
{\newenvironment{rep#1}[1]
{\def\rep@title{#2 \ref{##1}} \begin{rep@theorem}}%
 {\end{rep@theorem}}}
\makeatother
\newreptheorem{theorem}{Theorem}
\newreptheorem{lemma}{Lemma}
\newreptheorem{corollary}{Corollary}
\newreptheorem{proposition}{Proposition}

\newcommand*\samefootnotes[1][\value{footnote}]{\footnotemark[#1]}

\newcommand{\trace}{\ensuremath{\operatorname{trace}}}

\newcommand{\defn}{\ensuremath{: \, = }}

\newcommand{\Hil}{\ensuremath{\mathbb{H}}}
\newcommand{\ktil}{\ensuremath{\widetilde{k}}}

\newcommand{\var}{\ensuremath{\operatorname{var}}}

\newcommand{\order}{\ensuremath{\mathcal{O}}}

\title{Kernel Feature Selection via \\ Conditional Covariance
  Minimization}

\author{
Jianbo Chen\thanks{Equal contribution.}\\
University of California, Berkeley\\
\texttt{jianbochen@berkeley.edu}
\And
Mitchell Stern\samefootnotes\\
University of California, Berkeley\\
\texttt{mitchell@berkeley.edu}
\And
Martin J. Wainwright\\
University of California, Berkeley\\
\texttt{wainwrig@berkeley.edu}
\And
Michael I. Jordan\\
University of California, Berkeley\\
\texttt{jordan@berkeley.edu}\\
\and
}
\begin{document}

\maketitle

\addtocounter{footnote}{-1}

\begin{abstract}

We propose a method for feature selection that employs kernel-based measures of independence to find a subset of covariates that is maximally predictive of the response. Building on past work in kernel dimension reduction, we show how to perform feature selection via a constrained optimization problem involving the trace of the conditional covariance operator. We prove various consistency results for this procedure, and also demonstrate that our method compares favorably with other state-of-the-art algorithms on a variety of synthetic and real data sets.

\end{abstract}


\section{Introduction}

Feature selection is an important issue in statistical machine
learning, leading to both computational benefits (lower storage and
faster computation) and statistical benefits, including increased
model interpretability. With large data sets becoming ever more
prevalent, feature selection has seen widespread usage across a
variety of real-world tasks in recent years, including text
classification, gene selection from microarray data, and face
recognition~\cite{bolon2015recent,guyon2003introduction,li2016feature}. In
this work, we consider the supervised variant of feature selection,
which entails finding a subset of the input features that explains the
output well. This practice can reduce the computational expense of
downstream learning by removing features that are redundant or noisy,
while simultaneously providing insight into the data through the
features that remain.



Feature selection algorithms can generally be divided into two groups: those which are agnostic to the choice of learning algorithm, and those which attempt to find features that optimize the performance of a specific learning algorithm.%
\footnote{Feature selection algorithms that operate independently of the choice of predictor are referred to as filter methods. Algorithms tailored to specific predictors can be further divided into wrapper methods, which use learning algorithms to evaluate features based on their predictive power, and embedded methods, which combine feature selection and learning into a single problem~\cite{guyon2003introduction}.
}
%
Kernel methods have been successfully applied under
each of these paradigms in recent work; for instance, see the
papers~\cite{allen2013automatic,fukumizu2012gradient,jaganathan2011kernel,masaeli2010transformation, pmlr-v38-ren15,song2012feature,sun2014kernel,yamada2014high}.
Kernel feature selection methods have the advantage of capturing
nonlinear relationships between the features and the labels. Many
previous approaches are filter methods based on the Hilbert-Schmidt
Independence Criterion (HSIC), as proposed
by~\citet{gretton2005measuring} as a measure of dependence. For
instance, \citet{song2007supervised,song2012feature} proposed to
optimize HSIC with greedy algorithms on
features. \citet{masaeli2010transformation} proposed Hilbert-Schmidt
Feature Selection (HSFS), which optimizes HSIC with a continuous
relaxation. In later work, \citet{yamada2014high} proposed the
HSIC-LASSO, in which the dual augmented Lagrangian can be used to find
a global optimum. There are also wrapper methods and embedded methods
using kernels. Most of the methods add weights to features and
optimize the original kernelized loss function together with a penalty
on the
weights~\cite{allen2013automatic,cao2007feature,fukumizu2012gradient,gilad2004margin,grandvalet2002adaptive,weston2000feature,weston2003use}. For
example, \citet{cao2007feature} proposed margin-based algorithms for
SVMs to select features in the kernel space. Lastly,
\citet{allen2013automatic} proposed an embedded method suitable for
kernel SVMs and kernel ridge regression.

In this paper, we propose to use the trace of the conditional
covariance operator as a criterion for feature selection. We offer
theoretical motivation for this choice and show that our method can be
interpreted both as a filter method and as a wrapper method for a
certain class of learning algorithms. We also show that the empirical
estimate of the criterion is consistent as the sample size increases.
Finally, we conclude with an empirical demonstration that our
algorithm is comparable to or better than several other popular
feature selection algorithms on both synthetic and real-world tasks.



\section{Formulating feature selection}  

Let $\mathcal X\subset \mathbb{R}^d$ be the domain of covariates $X$,
and let $\mathcal Y$ be the domain of responses $Y$.  Given $n$
independent and identically distributed (i.i.d.) samples $\{
(x_i,y_i), \; i = 1,2,\dots,n \}$ generated from an unknown joint
distribution $P_{X,Y}$ together with an integer $m\leq d$, our goal is
to select $m$ of the $d$ total features $X_1,X_2,\dots,X_d$ which best
predict $Y$. Let $\mathcal{S}$ be the full set of features, and let
$\mathcal{T} \subseteq \mathcal{S}$ denote a subset of features. For
ease of notation, we identify $\mathcal{S}=\{X_1,\dots,X_d\}$ with
$[d]=\{1,\dots,d\}$, and also identify $X_\T$ with $\T$. We formulate
the problem of supervised feature selection from two perspectives
below. The first perspective motivates our algorithm as a filter
method. The second perspective offers an interpretation as a wrapper
method.


\subsection{From a dependence perspective}

Viewing the problem from the perspective of dependence, we would
ideally like to identify a subset of features $\mathcal{T}$ of size
$m$ such that the remaining features $\mathcal{S} \setminus
\mathcal{T}$ are conditionally independent of the responses given
$\mathcal{T}$. However, this may not be achievable when the number of
allowable features $m$ is small. We therefore quantify the extent of
the remaining conditional dependence using some metric $\mathcal{Q}$,
and aim to minimize $\mathcal{Q}$ over all subsets $\mathcal{T}$ of
the appropriate size. More formally, let $\mathcal{Q}: 2^{[d]} \to
[0,\infty)$ be a function mapping subsets of $[d]$ to the non-negative
  reals that satisfies the following properties:
\begin{itemize}
\item For a subset of features $\T$, we have $\mathcal{Q}(\T)=0$ if
  and only if $X_{\mathcal{S} \setminus \mathcal{T}}$ and $Y$ are
  conditionally independent given $X_\T$.
\item The function $\mathcal{Q}$ is non-increasing, meaning that
  $\mathcal{Q}(\mathcal{T}) \geq \mathcal{Q}(\mathcal{S})$ whenever
  $\mathcal{T} \subseteq \mathcal{S}$. Hence, the function
  $\mathcal{Q}$ achieves its minimum for the full feature set
  $\T=[d]$.
\end{itemize}
Given a fixed integer $m$, the problem of supervised feature selection
can then be posed as
\begin{align}
\min_{\T : |\T| = m} \mathcal{Q}(\T) .
\end{align}  
This formulation can be taken as a filter method for feature selection. 

\subsection{From a prediction perspective}

An alternative perspective aims at characterizing how well $X_\T$ can
predict $Y$ directly within the context of a specific learning
problem. Formally, we define the error of prediction as
\begin{align}
\mathcal E_{\mathcal F}(X) = \inf_{f\in \mathcal F} \mathbb{E}_{X,Y}
L(Y,f(X)),
\end{align}
where $\mathcal{F}$ is a class of functions from $\X$ to $\Y$, and $L$
is a \emph{loss function} specified by the user. For example, in a
univariate regression problem, the function class $\mathcal{F}$ might
be the set of all linear functions, and the loss function might be the
squared error $L(Y,f(X))=(Y-f(X))^2$.

We then hope to solve the following problem:
\begin{align*}
  \min_{\mathcal T: |\T| \leq m} \mathcal E_{\mathcal F}(X_\T) =
  \min_{\mathcal T: |\T| \leq m} \inf_{f\in \mathcal{F}_m}
  \mathbb{E}_{X,Y}L(Y,f(X_\T)),
\end{align*}
where $\mathcal F_m$ is a class of functions supported on $\R^m$. That
is, we aim to find the subset of $m$ features that minimizes the
prediction error. This formulation thus falls within the scope of
wrapper methods for feature selection.
\section{Conditional covariance operator}
The conditional covariance operator provides a measure of conditional
dependence for random variables. It was first proposed by
\citet{baker1973joint}, and was further studied and used for
sufficient dimension reduction by
\citet{fukumizu2004dimensionality,fukumizu2009kernel}. We provide a
brief overview of this operator and some of its key properties here.

Let $(\Hil_\X,k_\X)$ and $(\Hil_\Y,k_\Y)$ denote
reproducing kernel Hilbert spaces (RKHSs) of functions on spaces $\X$
and $\Y$, respectively. Also let $(X,Y)$ be a random vector on $\X
\times \Y$ with joint distribution $P_{X,Y}$. Assume the kernels
$k_\X$ and $k_\Y$ are bounded in expectation:
\begin{align}
\E_X[k_\X(X,X)]<\infty \quad\text{and}\quad \E_Y[k_\Y(Y,Y)]<\infty .
\end{align}
The cross-covariance operator associated with the pair $(X,Y)$ is
the mapping $\Sigma_{YX}:\Hil_\X\to\Hil_\Y$ defined by
the relations
\begin{align}
\langle g,\Sigma_{YX}f\rangle_{\Hil_\Y} = \mathbb
E_{X,Y}[(f(X)-\mathbb E_X[f(X)])(g(Y)-\E_Y[g(Y)])] \quad \mbox{for
  all $f \in \Hil_X$ and $g \in \Hil_Y$.}
\end{align}
\citet{baker1973joint} showed there exists a unique bounded operator
$V_{YX}$ such that
\begin{subequations}
\begin{align}
\Sigma_{YX} = \Sigma_{YY}^{1/2}V_{YX}\Sigma_{XX}^{1/2}.
\end{align}
The conditional covariance operator is then defined as 
\begin{align}
\Sigma_{YY|X}=\Sigma_{YY} - \Sigma_{YY}^{1/2}V_{YX}V_{XY}\Sigma_{YY}^{1/2}.
\end{align}
\end{subequations}
Among other results,
\citet{fukumizu2004dimensionality,fukumizu2009kernel} showed that the
conditional covariance operator captures the conditional variance of
$Y$ given $X$. More precisely, if the sum $\Hil_\X + \R$ is dense in
$L^2(P_X)$, where $L^2(P_X)$ is the space of all square-integrable
functions on $\X$, then we have
\begin{align}
  \label{equation:conditonal variance}
\langle g,\Sigma_{YY|X}g\rangle_{\Hil_\Y} = \E_X[\var_{Y|X}[g(Y)|X]]
\quad \mbox{for any $g\in \Hil_\Y$.}
\end{align}
From Proposition 2 in the paper~\cite{fukumizu2009kernel}, we also
know the residual error of $g(Y)$ with $g\in\Hil_\Y$ can be
characterized by the conditional covariance operator. More formally,
for any $g\in\Hil_\Y$, we have
\begin{align}
  \label{prop:residual}
\langle g,\Sigma_{YY|X}g\rangle_{\Hil_\Y} = \inf_{f\in\mathcal
  H_\X}\E_{X,Y} ((g(Y)-\E_Y[g(Y)])-(f(X)-\E_X[f(X)]))^2.
\end{align}


\section{Proposed method}  

In this section, we describe our method for feature selection, which we call conditional covariance minimization (CCM).


Let $(H_1,k_1)$ denote an RKHS supported on $\X \subset \R^d$. Let $\T
\subseteq [d]$ be a subset of features with cardinality $m \le d$, and
for all $x \in \R^d$, take $x^\T \in \R^d$ to be the vector with
components $x^\T_i = x_i$ if $i \in \T$ or $0$ otherwise. We define
the kernel $k_1^\T$ by $k_1^\T(x,\tilde x)=k_1(x^\T,\tilde x^\T)$ for
all $x,\tilde x\in\X$. Suppose further that the kernel $k_1$ is
permutation-invariant. That is, for any $x,\tilde x\in\X$ and
permutation $\pi$, denoting $(x_{\pi(1)},\dots,x_{\pi(d)})$ as
$x_\pi$, we have $k_1(x,\tilde x)=k_1(x_{\pi},\tilde x_\pi)$. (Note
that this property holds for many common kernels, including the
linear, polynomial, Gaussian, and Laplacian kernels.) Then for every
$\T$ of cardinality $m$, $k_1^\T$ generates the same RKHS supported on
$\R^m$. We call this RKHS $(\tilde H_1,\ktil_1)$. We will show the
trace of the conditional covariance operator $\trace(\Sigma_{YY|X})$
can be interpreted as a dependence measure, as long as the RKHS $H_1$
is large enough.

We say that an RKHS $(H,k)$ is characteristic if the map $P\to
\E_P[k(X,\cdot)]\in H$ is one-to-one. If $k$ is bounded, this is
equivalent to saying that $H+\R$ is dense in $L^2(P)$ for any
probability measure $P$~\cite{fukumizu2009kernel}. We have the
following lemma, whose proof is given in the appendix:
\begin{lemma}
\label{lemma:characteristic}
If $k_1$ is bounded and characteristic, then $\ktil_1$ is also
characteristic.
\end{lemma}

Let $(H_2,k_2)$ denote an RKHS supported on $\Y$. Based on the above lemma, we have the following theorem, which is a
parallel version of Theorem 4 in \cite{fukumizu2009kernel}:
\begin{theorem}
\label{theorem:general}
If $(H_1,k_1)$ and $(H_2,k_2)$ are characteristic, we have
$\Sigma_{YY|X} \preceq \Sigma_{YY|X_\T}$ with equality holding if and
only if $Y\independent X|X_\T$.
\end{theorem}  

The proof is postponed to the appendix.

With this generic result in place, we now narrow our focus to problems with univariate responses, including univariate regression, binary classification and multi-class classification. In the case of regression, we assume $H_2$ is supported on $\R$, and we take $k_2$ to be the linear kernel:
\begin{align}
k_2(y,\tilde y) = y\tilde y
\end{align}
for all $y,\tilde y\in \R$. For binary or multi-class classification, we take $k_2$ to be the Kronecker delta function:
\begin{align}
k_2(y,\tilde y) = \delta(y,\tilde y) = \begin{cases} 1 & \text{if } y = \tilde{y} , \\ 0 & \text{otherwise} . \end{cases}
\end{align}
This can be equivalently interpreted as a linear kernel $k(y,\tilde y)=\langle y,\tilde y\rangle$ assuming a one-hot encoding of $Y$, namely that $\mathcal Y = \{ y \in \{0,1\}^k : \sum_i y_i = 1 \} \subset \mathbb{R}^k$, where $k$ is the number of classes. 


When $\Y$ is $\R$ or $\{ y \in \{0,1\}^k : \sum_i y_i = 1 \} \subset \mathbb{R}^k$, we obtain the following corollary of Theorem~\ref{theorem:general}:

\begin{corollary}\label{cor:univariate}
If $(H_1,k_1)$ is characteristic, $\Y$ is $\R$ or $\{ y \in \{0,1\}^k : \sum_i y_i = 1 \} \subset \mathbb{R}^k$, and $(H_2,k_2)$ includes the identity function on $\Y$, then we have $\mathrm{Tr}(\Sigma_{YY|X})\leq\mathrm{Tr}(\Sigma_{YY|X_\T})$ for any subset $\T$ of features. Moreover, the equality $\mathrm{Tr}(\Sigma_{YY|X})=\mathrm{Tr}(\Sigma_{YY|X_\T})$ holds if and only if $Y\independent X|X_\T$.
\end{corollary}

Hence, in the univariate case, the problem of supervised feature selection reduces to minimizing the trace of the conditional covariance operator over subsets of features with controlled cardinality:
\begin{align}
\min_{\mathcal T:|\T|= m} \mathcal{Q}(\T) := \mathrm{Tr}(\Sigma_{YY|X_\T}).
\end{align} 
In the regression setting, Equation (\ref{prop:residual}) implies the residual error of regression can also be characterized by the trace of the conditional covariance operator when using the linear kernel on $\Y$. More formally, we have the following observation:
\begin{corollary}
Let $\Sigma_{YY|X_\T}$ denote the conditional covariance operator of $(X_\T, Y)$ in $(\tilde H_1,\ktil_1)$. Define the space of functions $\mathcal F_m$ from $\R^m$ to $\Y$ as 
\begin{align}
\mathcal F_m = \tilde H_1+\R: = \{f+c: f\in \tilde H_1,c\in \R\}.
\end{align}
Then we have 
\begin{align}\label{equation:population}
\mathrm{Tr}(\Sigma_{YY|X_\T}) = \mathcal E_{\mathcal F_m}(X_\T) =
\inf_{f\in\mathcal F_m}\E_{X,Y}(Y-f(X_\T))^2.
\end{align}
\end{corollary}

Given the fact that the trace of the conditional covariance operator
can characterize the dependence and the prediction error in
regression, we will use the empirical estimate of it as our
objective. Given $n$ samples $\{(x_1,y_1),\dots,(x_n,y_n)\}$, the
empirical estimate is given by \cite{fukumizu2009kernel}:
\begin{align*}
\trace(\hat\Sigma_{YY|X_\T}^{(n)})&:=\trace[\hat \Sigma_{YY}^{(n)} -
  \hat \Sigma_{YX_\T}^{(n)}(\hat \Sigma_{X_\T
    X_\T}^{(n)}+\varepsilon_n I)^{-1}\hat \Sigma_{X_\T
    Y}^{(n)}]
  \\ &=\varepsilon_n \trace[G_Y(G_{X_\T}+n\varepsilon_nI_n)^{-1}],
\end{align*}
where $\hat\Sigma^{\T(n)}_{YX},\hat\Sigma_{X_\T X}^{\T(n)}$ and
$\hat\Sigma_{YY}^{(n)}$ are the covariance operators defined with
respect to the empirical distribution and $G_{X_\T}$ and $G_Y$ are the
centralized kernel matrices, respectively. Concretely, we define
\begin{align*}
G_{X_\T} & \defn (I_n - \frac{1}{n} \mathbbm{1}
\mathbbm{1}^\T)K_{X_\T} (I_n - \frac{1}{n}\mathbbm{1}\mathbbm{1}^T)
\quad \mbox{and} \quad G_Y \defn (I_n - \frac{1}{n}
\mathbbm{1}\mathbbm{1}^T)K_Y
(I_n-\frac{1}{n}\mathbbm{1}\mathbbm{1}^T).
\end{align*}
The $(i,j)$th entry of the kernel matrix $K_{X_\T}$ is $\tilde
k_1(x_\T^i,x_\T^j)$, with $x_\T^i$ denoting the $i$th sample with only
features in $\T$. As the kernel $k_2$ on the space of responses is
linear, we have $K_Y=\bY\bY^T$, where $\bY$ is the $n\times k$ matrix
with each row being a sample response. Without loss of generality, we
assume each column of $\bY$ is zero-mean, so that
$G_Y=K_Y=\bY\bY^T$. Our objective then becomes:
\begin{align}
\trace[G_Y(G_{X_\T}+n\varepsilon_nI_n)^{-1}] =
\trace[\bY\bY^T(G_{X_\T}+n\varepsilon_nI_n)^{-1}]=
\trace[\bY^T(G_{X_\T}+n\varepsilon_nI_n)^{-1}\bY].
\end{align}
For simplicity, we only consider univariate regression and binary
classification where $k=1$, but our discussion carries over to the
multi-class setting with minimal modification. The objective becomes
\begin{align}
\min_{|\T|= m} \hat{\mathcal Q}^{(n)} (\T):=
\by^T(G_{X_\T}+n\varepsilon_nI_n)^{-1}\by,
\label{equation:obj1}
\end{align}
where $\by=(y_1,\dots,y_n)^T$ is an $n$-dimensional vector. We show
the global optimal of the problem~\eqref{equation:obj1} is
consistent. More formally, we have the following theorem:
\begin{theorem}[Feature Selection Consistency]
  \label{theorem:consistency}
Let the set $A=\text{argmin}_{|\T|\leq m}\mathcal Q(\mathcal T)$ be
the set of all the optimal solutions to (\ref{equation:population})
and $\hat T^{(n)}\in \text{argmin}_{|T|\leq m}\hat{\mathcal Q}^{(n)}
(\T)$ be a global optimal of (\ref{equation:obj1}). If
$\varepsilon_n\to 0$ and $\varepsilon_n n\to\infty$ as $n\to\infty$,
we have
\begin{align}
\label{consistency}
P(\hat T^{(n)}\in A)\to 1.
\end{align} 
\end{theorem}
Our proof is provided in the appendix. A comparable result is given in
\citet{fukumizu2009kernel} for the consistency of their dimension
reduction estimator, but as our minimization takes place over a finite
set, our proof is considerably simpler.


\section{Optimization}
\label{sec:optimization}

Finding a global optimum for \eqref{equation:obj1} is NP-hard for
generic kernels \cite{weston2003use}, and exhaustive search is
computationally intractable if the number of features is large. We
therefore approximate the problem of interest via continuous
relaxation, as has previously been done in past work on feature
selection~\cite{bradley1998feature,weston2000feature,weston2003use}.


\subsection{Initial relaxation}
\label{subsec:initial-relaxation}

We begin by introducing a binary vector $w \in\{0,1\}^d$ to indicate
which features are active. This allows us to rephrase the optimization
problem from \eqref{equation:obj1} as
\begin{align}
\begin{aligned}
& \min_{w} & & \by^T(G_{w\odot X}+n\varepsilon_n I_n)^{-1}\by \\ &
  \text{subject to} & & w_i \in \{0, 1\}, \; i = 1, \dots, d, \\ & & &
  \mathbbm{1}^T w = m ,
\end{aligned}
\label{equation:obj2}
\end{align} 
where $\odot$ denotes the Hadamard product between two vectors and
$G_{w\odot X}$ is the centralized version of the kernel matrix $K_{w\odot X}$ with
$(K_{w\odot X})_{ij}=k_1(w\odot x_i,w\odot x_j)$.

We then approximate the problem~\eqref{equation:obj2} by relaxing the domain of $w$ to the unit hypercube $[0,1]^d$ and replacing the equality constraint with an inequality constraint:
\begin{align}
\begin{aligned}
& \underset{w}{\text{min}} & & \by^T(G_{w\odot
    X}+n\varepsilon_nI_n)^{-1}\by \\ & \text{subject to} & & 0\leq w_i
  \leq 1, \; i = 1, \ldots, d , \\ & & & \mathbbm{1}^T w \leq m .
\end{aligned}
\label{obj3}
\end{align}
This objective can be optimized using projected gradient descent, and
represents our first tractable approximation. A solution to the
relaxed problem is converted back into a solution for the original
problem by setting the $m$ largest values of $w$ to $1$ and remaining
values to $0$. We initialize $w$ to the uniform vector
$(m/d)[1,1,\dots,1]^T$ in order to avoid the corners of the constraint
set during the early stages of optimization.


\subsection{Computational issues}
\label{subsec:computational-issues}

The optimization problem can be approximated and manipulated in a
number of ways so as to reduce computational complexity. We
discuss a few such options below.

\paragraph{Removing the inequality constrant.}

The hard constraint $\mathbbm{1}^T w \leq m$ requires a nontrivial
projection step, such as the one detailed in
\citet{duchi2008efficient}. We can instead replace it with a soft
constraint and move it to the objective. Letting $\lambda_1 \ge 0$ be
a hyperparameter, this gives rise to the modified problem
\begin{align}
\begin{aligned}
& \underset{w}{\text{min}} & & \by^T(G_{w\odot X}+ n
  \varepsilon_nI_n)^{-1} \by +
\lambda_1 (\mathbbm{1}^T w - m) \\
& \text{subject to} & & 0\leq w_i \leq 1, \; i = 1, \ldots, d .
\end{aligned}
\end{align}


\paragraph{Removing the matrix inverse.}

The matrix inverse in the objective function is an expensive
operation. In light of this, we first define an auxiliary variable
$\alpha \in \R^n$, add the equality constraint $\alpha = (G_{w\odot
  X}+n\varepsilon_n I_n)^{-1}\by$, and rewrite the objective as
$\alpha^T \by$. We then note that we may multiply both sides of the
constraint by the centered kernel matrix to obtain the relation
$(G_{w\odot X}+n\varepsilon_n I_n) \alpha = \by$. Letting $\lambda_2
\ge 0$ be a hyperparameter, we finally replace this relation by a soft
$\ell_2$ constraint to obtain
\begin{align}
\begin{aligned}
& \underset{w,\alpha}{\text{min}} & & \alpha^T\by + \lambda_2 \|(G_{w\odot X}+n\varepsilon_nI_n)\alpha - \by\|_2^2 \\
& \text{subject to} & & 0\leq w_i \leq 1, \; i = 1, \ldots, d , \\
& & & \mathbbm{1}^T w \leq m .
\end{aligned}
\end{align}



\paragraph{Using a kernel approximation.}

\citet{rahimi07random} propose a method for approximating kernel
evaluations by inner products of random feature vectors, so that
$k(x,\tilde x) \approx z(x)^T z(\tilde x)$ for a random map $z$
depending on the choice of kernel $k$. Let $K_w \approx U_w U_w^T$ be
such a decomposition, where $U_w\in\R^{n\times D}$ for some
$D<n$. Then, defining $V_w = (I - \mathbbm{1}\mathbbm{1}^T / n) U_w$,
we similarly have that the centered kernel matrix can be written as
$G_w \approx V_w V_w^T$. By the Woodbury matrix identity, we may write
\begin{align}
\begin{aligned}
(G_{w\odot X} + n \varepsilon_n I_n)^{-1} & \approx
  \frac{1}{\varepsilon_n n} I - \frac{1}{\varepsilon_n^2 n^2} V_w (I_D
  + \frac{1}{\varepsilon_n n} V_w^T V_w)^{-1} V_w^T \\ & =
  \frac{1}{\varepsilon_n n} (I - V_w (V_w^T V_w + \varepsilon_n n
  I_D)^{-1} V_w^T) .
\end{aligned}
\end{align}
Substituting this into our objective function, scaling by $\epsilon_n
n$, and removing the constant term $\by^T \by$ resulting from the
identity matrix gives a new approximate optimization problem.
This modification reduces the complexity of each optimization step
from $\order(n^2 d + n^3)$ to $\order(n^2 D + D^3 + nDd)$.

\paragraph{Choice of formulation.}

We remark that each of the three approximations beyond the initial
relaxation may be independently used or omitted, allowing for a number
of possible objectives and constraint sets. We explore some of these
configurations in the experimental section below.

\section{Experiments}

In this section, we evaluate our approach (CCM) on both synthetic and
real-world data sets. We compare with several strong existing
algorithms, including recursive feature elimination
(RFE)~\cite{guyon2002gene}, Minimum Redundancy Maximum Relevance
(mRMR)~\cite{peng2005feature},
BAHSIC~\cite{song2007supervised,song2012feature}, and filter methods
using mutual information (MI) and Pearson's correlation (PC). We use
the author's implementation for
BAHSIC\footnote{\url{http://www.cc.gatech.edu/~lsong/code.html}} and
use the Scikit-learn~\cite{scikit-learn} and
Scikit-feature~\cite{li2016feature} packages for the rest of the
algorithms. The code for our approach is publicly available at \url{https://github.com/Jianbo-Lab/CCM}.


\subsection{Synthetic data}


We begin with experiments on the following synthetic data sets:
\begin{itemize} 
\item Binary classification (\citet{friedman2001elements}). Given
  $Y=-1$, $(X_1,\dots,X_{10}) \sim N(0,I_{10})$. Given $Y=1$, $X_1$
  through $X_4$ are standard normal conditioned on $9\leq \sum_{j=1}^4
  X_j^2\leq 16$, and $(X_5,\dots,X_{10}) \sim N(0,I_6)$. 
\item 3-dimensional XOR as 4-way classification. Consider the 8
  corners of the 3-dimensional hypercube $(v_1, v_2, v_3) \in
  \{-1,1\}^3$, and group them by the tuples $(v_1 v_3, v_2 v_3)$,
  leaving 4 sets of vectors paired with their negations $\{v^{(i)},
  -v^{(i)}\}$. Given a class $i$, a point is generated from the
  mixture distribution $(1/2)N(v^{(i)}, 0.5 I_3) + (1/2)N(-v^{(i)},
  0.5 I_3)$. Each example additionally has 7 standard normal noise
  features for a total of 10 dimensions.

\item Additive nonlinear regression:
  $Y=-2\sin(2X_1)+\max(X_2,0)+X_3+\exp(-X_4)+\varepsilon$, where
  $(X_1,\dots,X_{10})\sim N(0,I_{10})$ and $\varepsilon\sim N(0,1)$.
%
\end{itemize}

The first data set represents a standard nonlinear binary classification task.
The second data set is a multi-class classification task where each
feature is independent of $Y$ by itself but a combination of three features
has a joint effect on $Y$. The third data set arises from an additive
model for nonlinear regression.


\begin{figure}[t]
\centering
\includegraphics[width=0.33\linewidth]{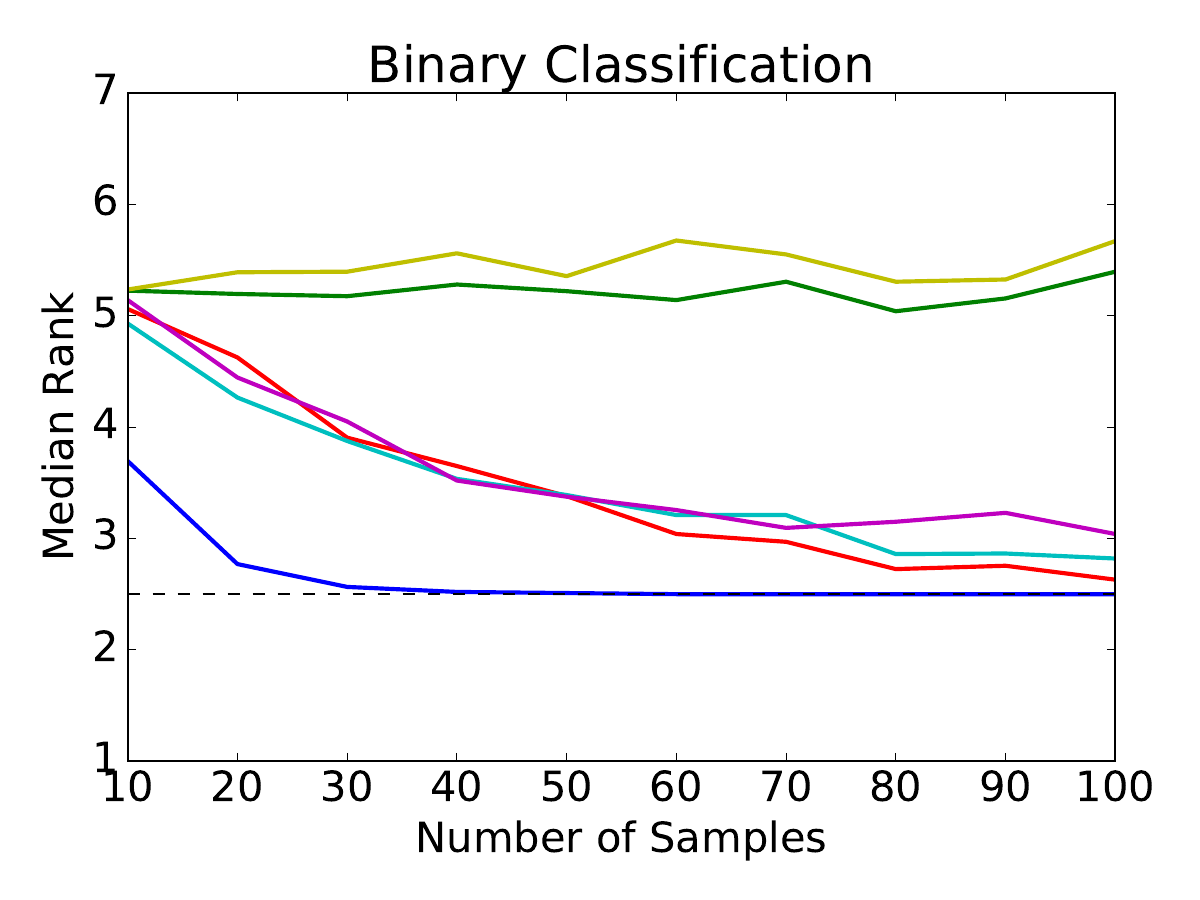}%
\includegraphics[width=0.33\linewidth]{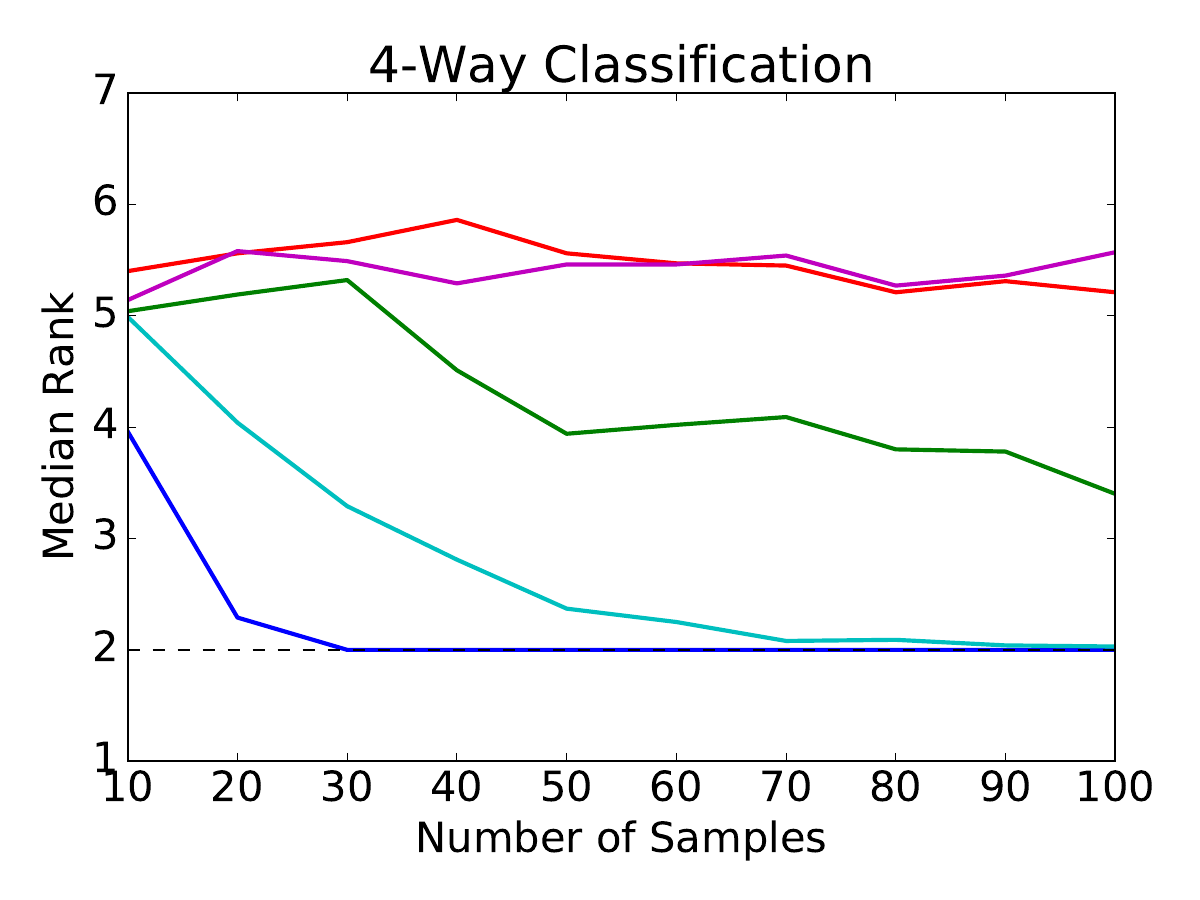}%
\includegraphics[width=0.33\linewidth]{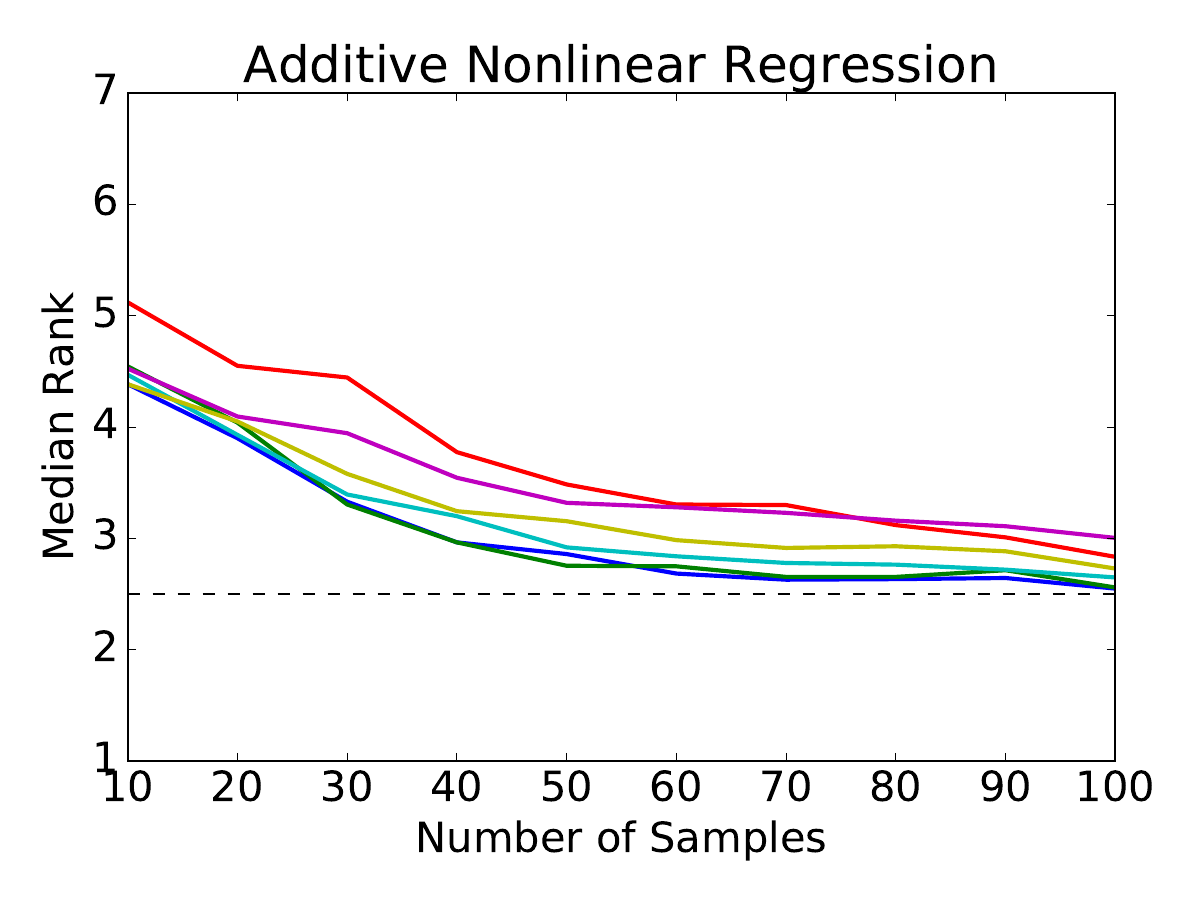} \\
\includegraphics[height=1.25em]{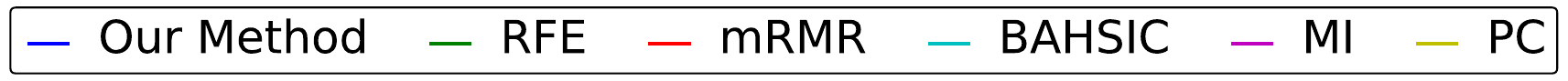}
\caption{The above plots show the median rank ($y$-axis) of the true
  features as a function of sample size ($x$-axis) for the simulated
  data sets. Lower median ranks are better. The dotted line indicates
  the optimal median rank.}
\label{fig:results-synthetic} 
\end{figure}

Each data set has $d=10$ dimensions in total, but only $m=3$ or $4$
true features. Since the identity of these features is known, we can
evaluate the performance of a feature selection algorithm by computing
the median rank it assigns to the real features, with lower median
ranks indicating better performance. Given enough samples, we would
expect this value to come close to the optimal lower bound of
$(m+1)/2$.

Our experimental setup is as follows. We generate 10 independent
copies of each data set with sample sizes ranging from 10 to 100, and
record the median ranks assigned to the true features by each
algorithm. This process is repeated a total of 100 times, and the
results are averaged across trials. For kernel-based methods, we use a
Gaussian kernel $k(x,\tilde x) = \exp(-\|x-\tilde{x}\|^2/(2\sigma^2))$
on $X$ and a linear kernel $k(y,\tilde y) = y^T \tilde y$ on $Y$. We
take $\sigma$ to be the median pairwise distance between samples
scaled by $1/\sqrt{2}$. Since the number of true features is known, we
provide this as an input to algorithms that require it.

Our initial experiments use the basic version of our algorithm from
Section~\ref{subsec:initial-relaxation}. When the number of desired
features $m$ is fixed, only the regularization parameter $\varepsilon$
needs to be chosen. We use $\varepsilon = 0.001$ for the
classification tasks and $\varepsilon = 0.1$ for the regression task,
selecting these values from $\{0.001, 0.01, 0.1\}$ using
cross-validation. Our results are shown in
Figure~\ref{fig:results-synthetic}.

On the binary and 4-way classification tasks, our method outperforms
all other algorithms, succeeding in identifying the true features
using fewer than 50 samples where others require close to 100 or even
fail to converge.
On the additive nonlinear model, several algorithms perform well, and
our method is on par with the best of them across all sample sizes.

These experiments show that our algorithm is comparable to or better
than several widely-used feature selection techniques on a selection
of synthetic tasks, and is adept at capturing several kinds of
nonlinear relationships between the covariates and the responses. When
compared in particular to its closest relative BAHSIC, a
backward-elimination algorithm based on the Hilbert–Schmidt
independence criterion, we see that our algorithm often produces
higher quality results with fewer samples, and even succeeds in the
non-additive problem where BAHSIC fails to converge.

We also rerun these experiments separately for each of the first two approximations described in Section~\ref{subsec:computational-issues} above, selecting $\lambda_1$ from $\{0.001, 0.01, 0.1\}$ and $\lambda_2$ from $\{1, 10, 100\}$ using cross-validation. We find that comparable results can be attained with either approximate objective, but note that the algorithm is more robust to changes in $\lambda_1$ than $\lambda_2$.


\subsection{Real-world data}

In the previous section, we found that our method for feature
selection excelled in identifying nonlinear relationships on a variety
of synthetic data sets. We now turn our attention to a collection of
real-word tasks, studying the performance of our method and other
nonlinear approaches when used in conjunction with a kernel SVM for
downstream classification.


We carry out experiments on 12 standard benchmark tasks from the ASU
feature selection website~\cite{li2016feature} and the UCI
repository~\cite{Lichman:2013}. A summary of our data sets is provided
in Table~\ref{tab:real-data-summary}. The data sets are drawn from
several domains including gene data, image data, and voice data, and
span both the low-dimensional and high-dimensional regimes.

\begin{figure}[t]
\centering
\includegraphics[width=0.25\linewidth]{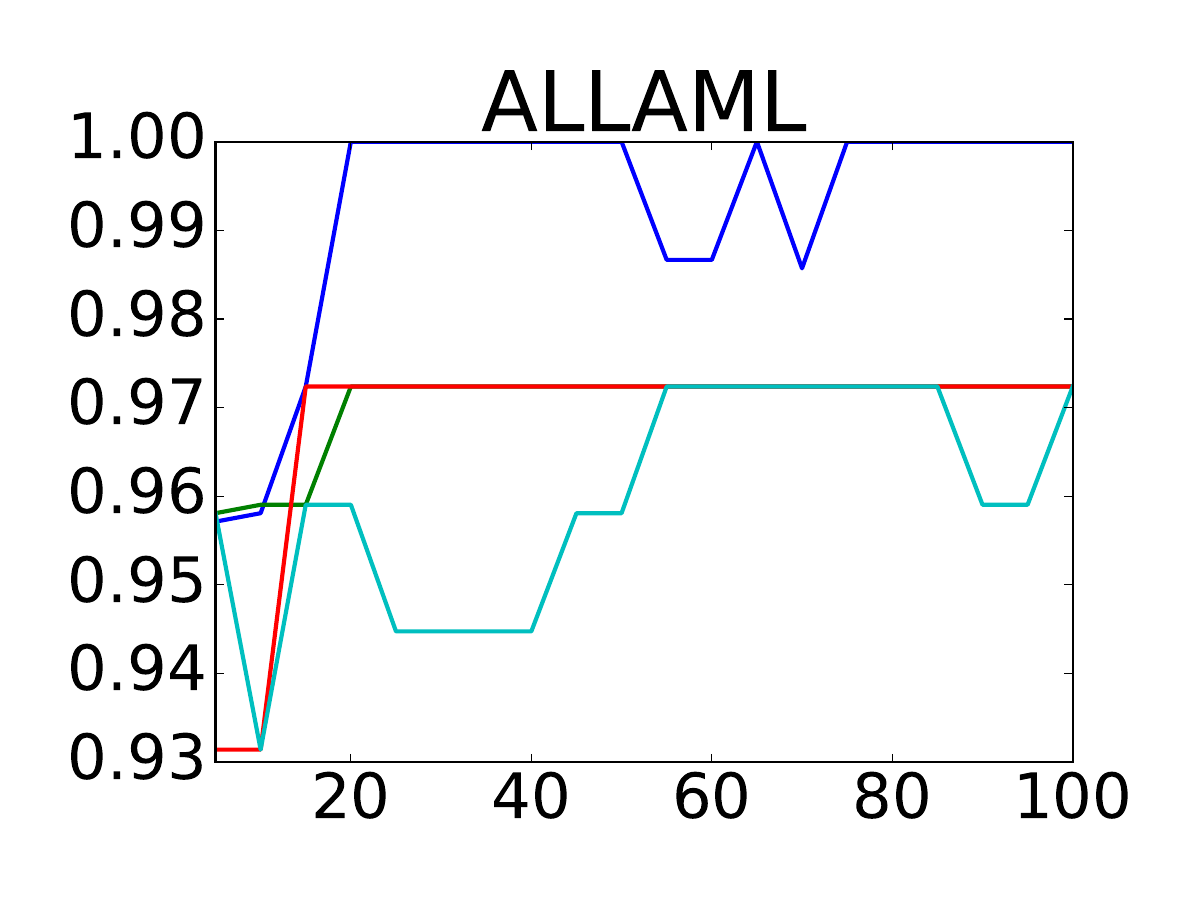}
\includegraphics[width=0.25\linewidth]{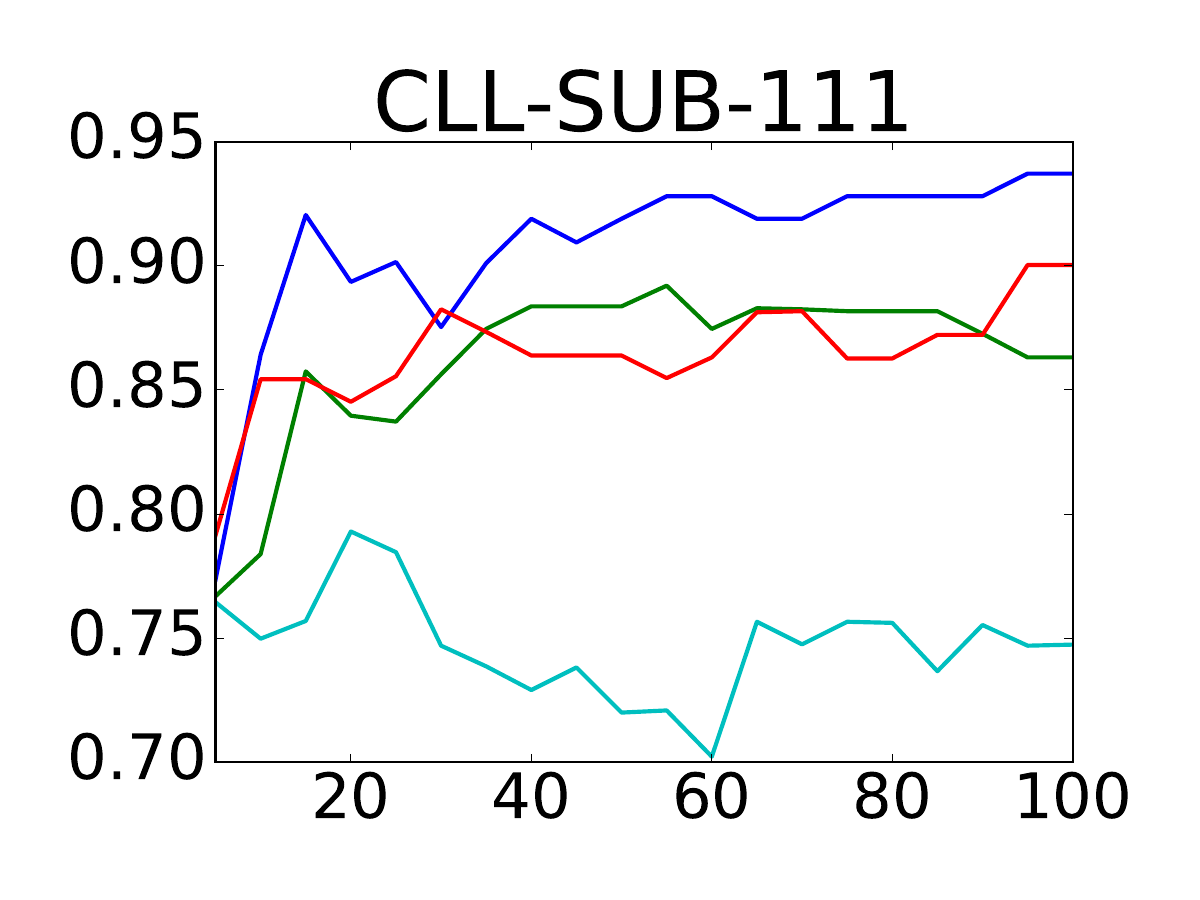}
\includegraphics[width=0.25\linewidth]{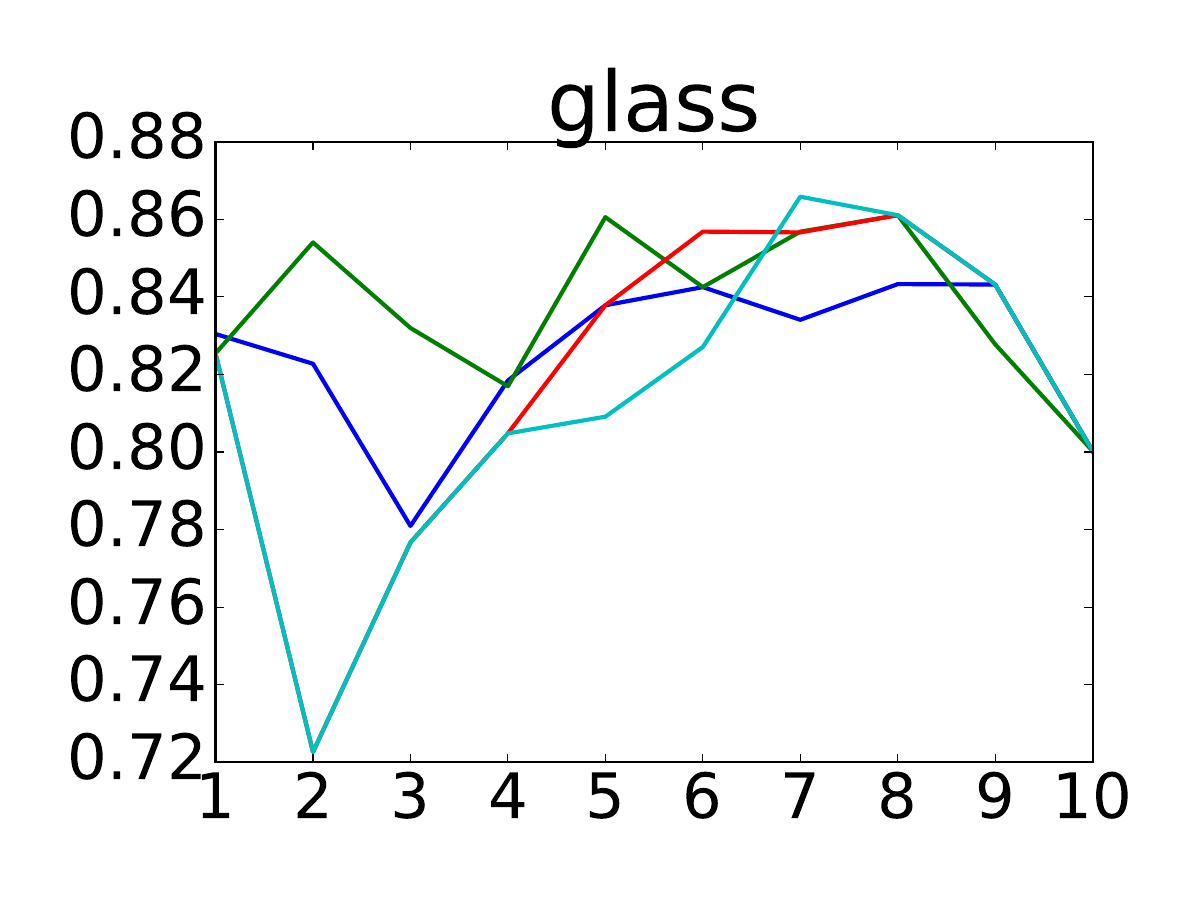}%
\includegraphics[width=0.25\linewidth]{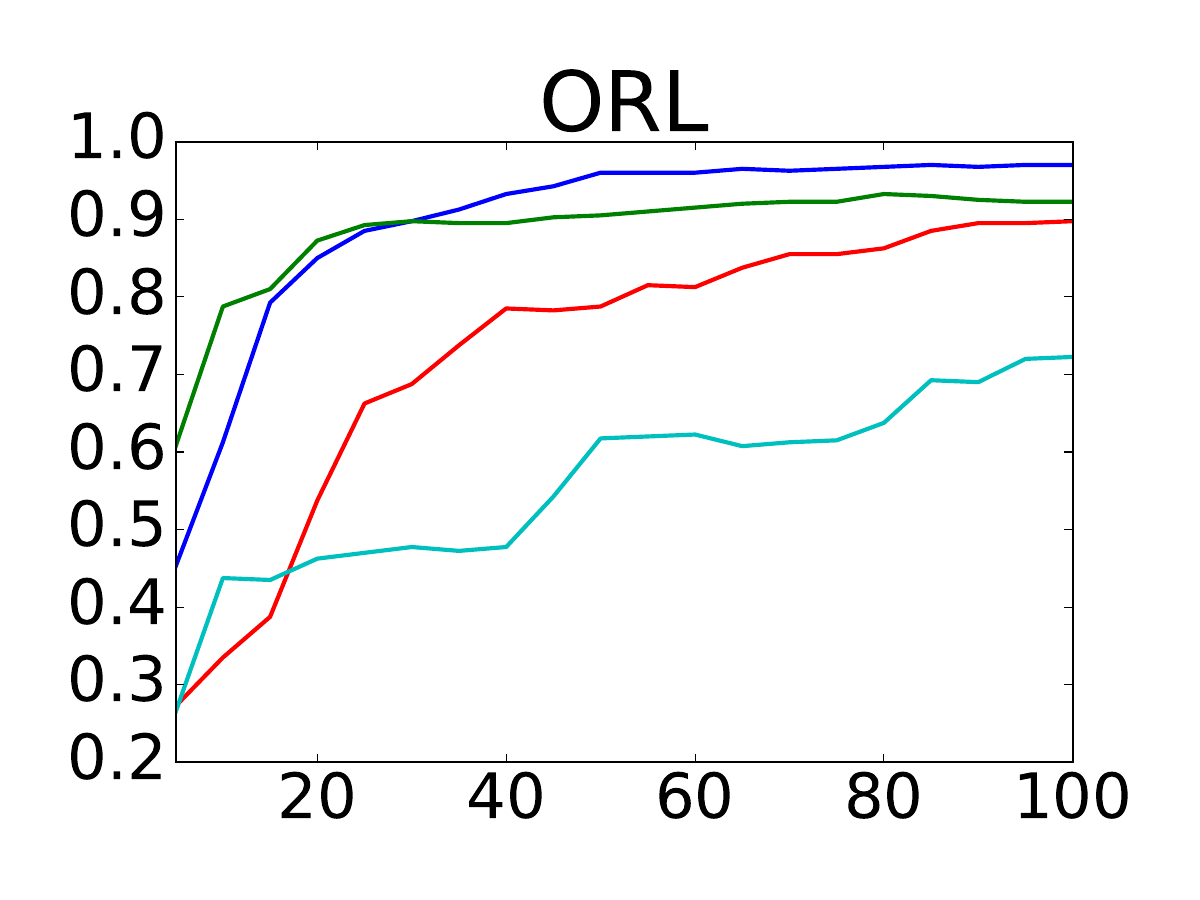} \\ 
\includegraphics[width=0.25\linewidth]{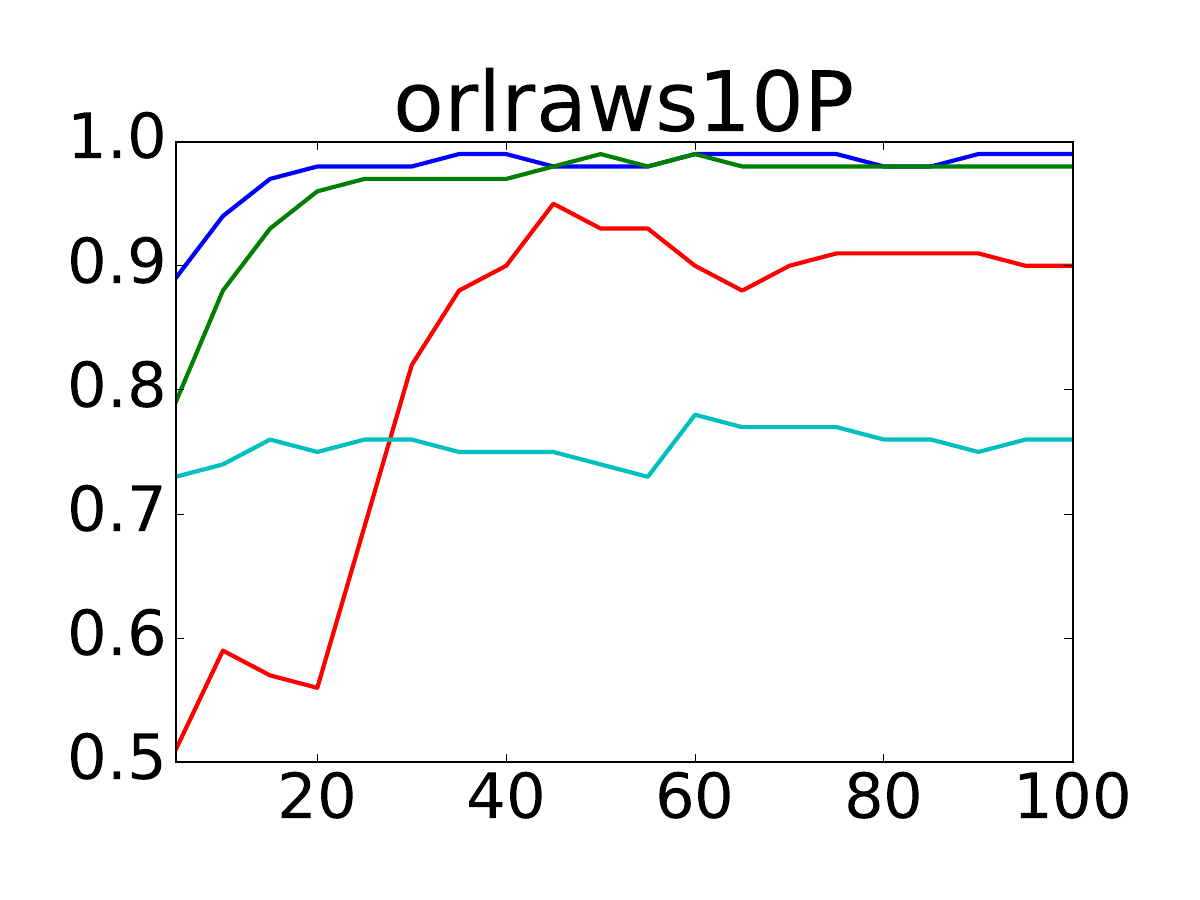}
\includegraphics[width=0.25\linewidth]{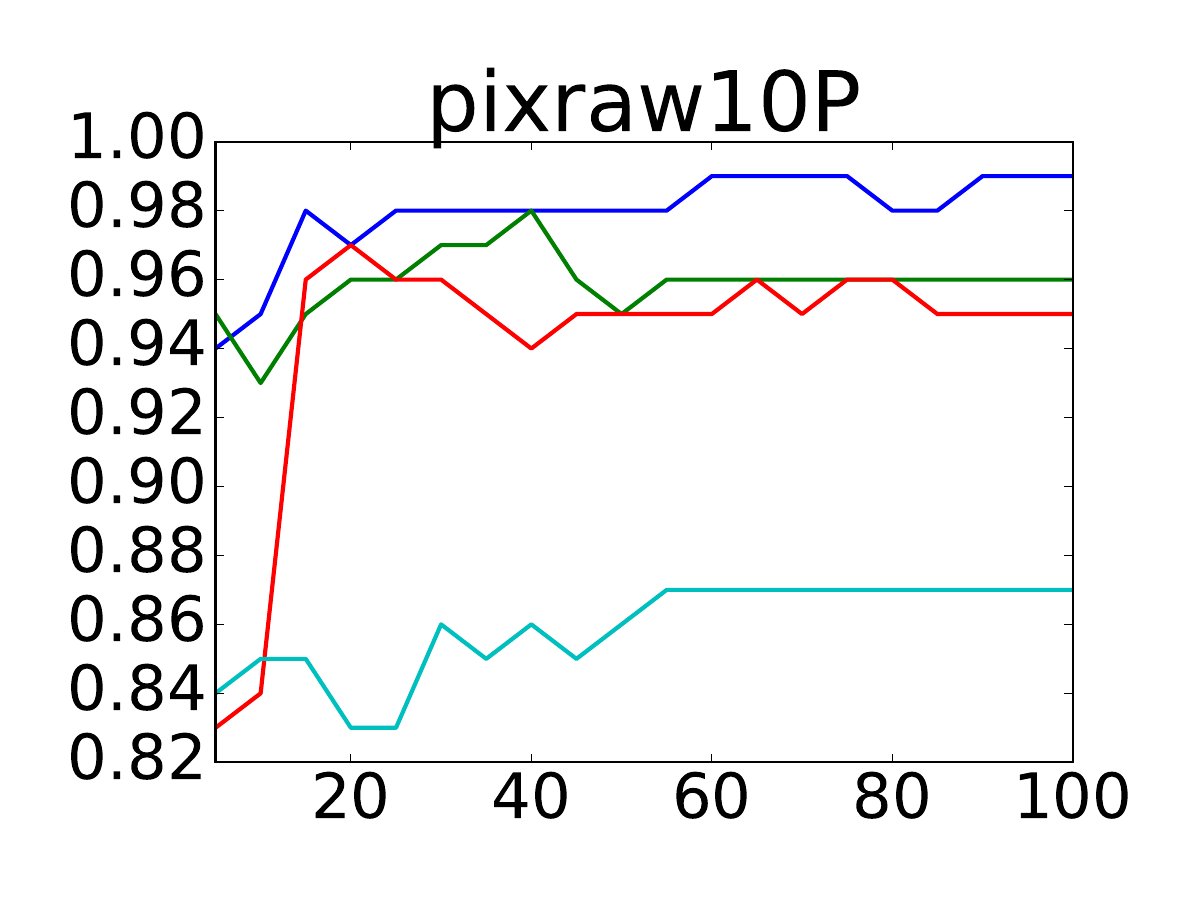}
\includegraphics[width=0.25\linewidth]{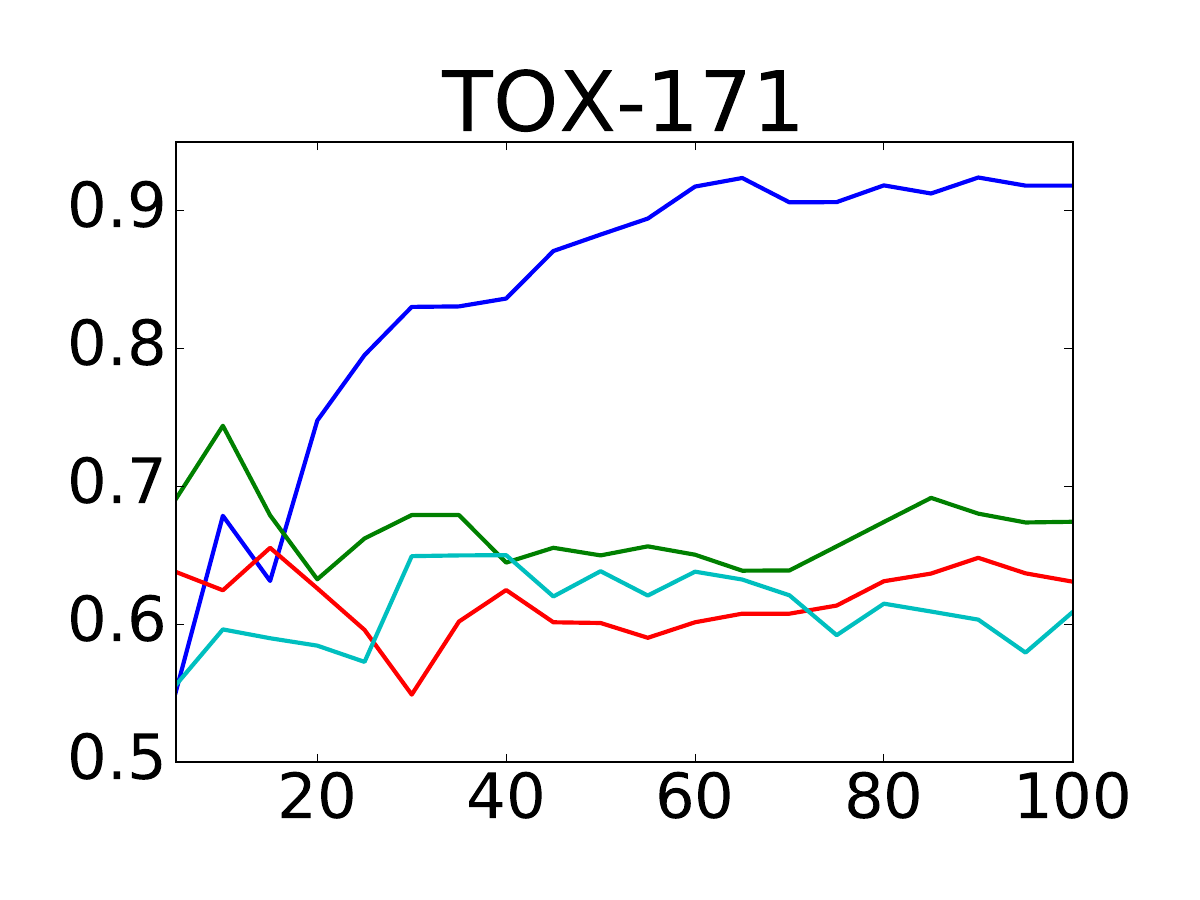}
\includegraphics[width=0.25\linewidth]{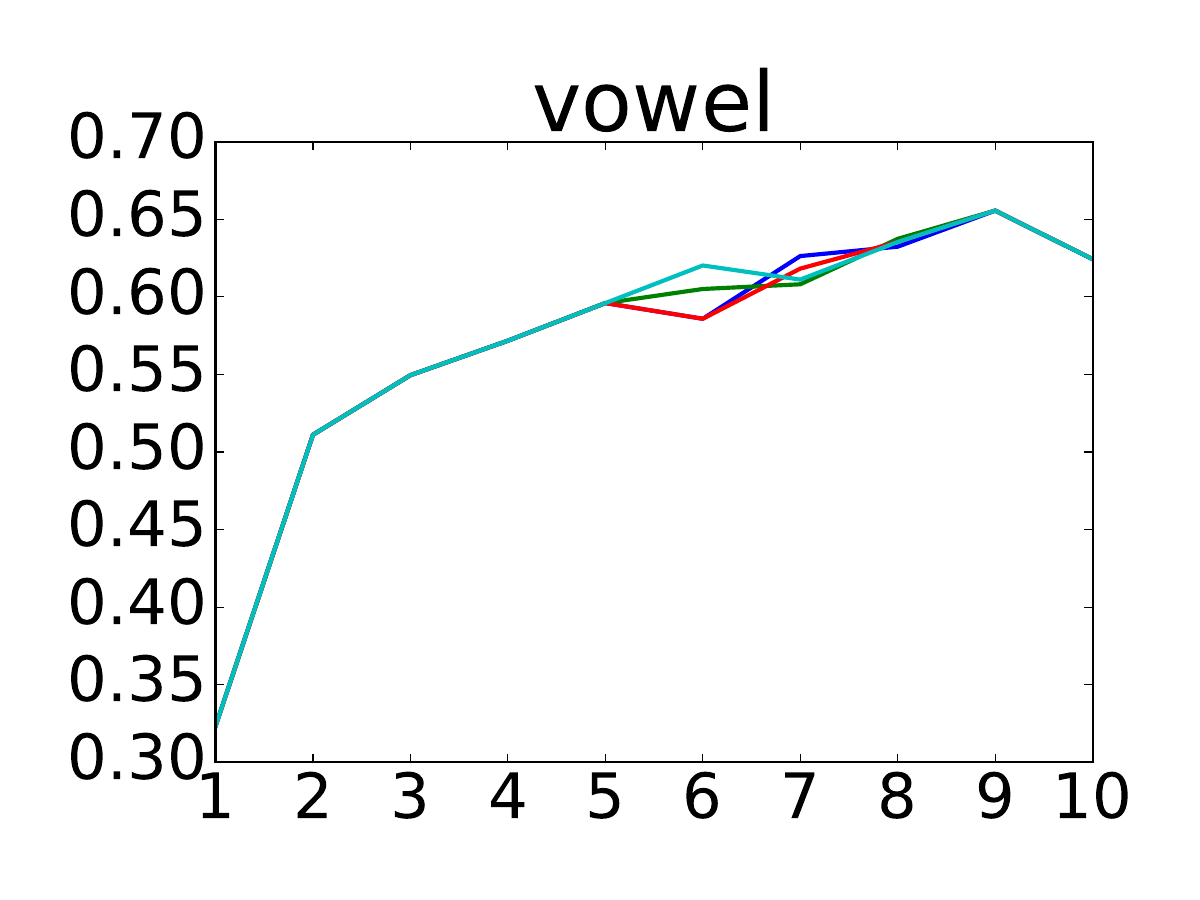} \\
\includegraphics[width=0.25\linewidth]{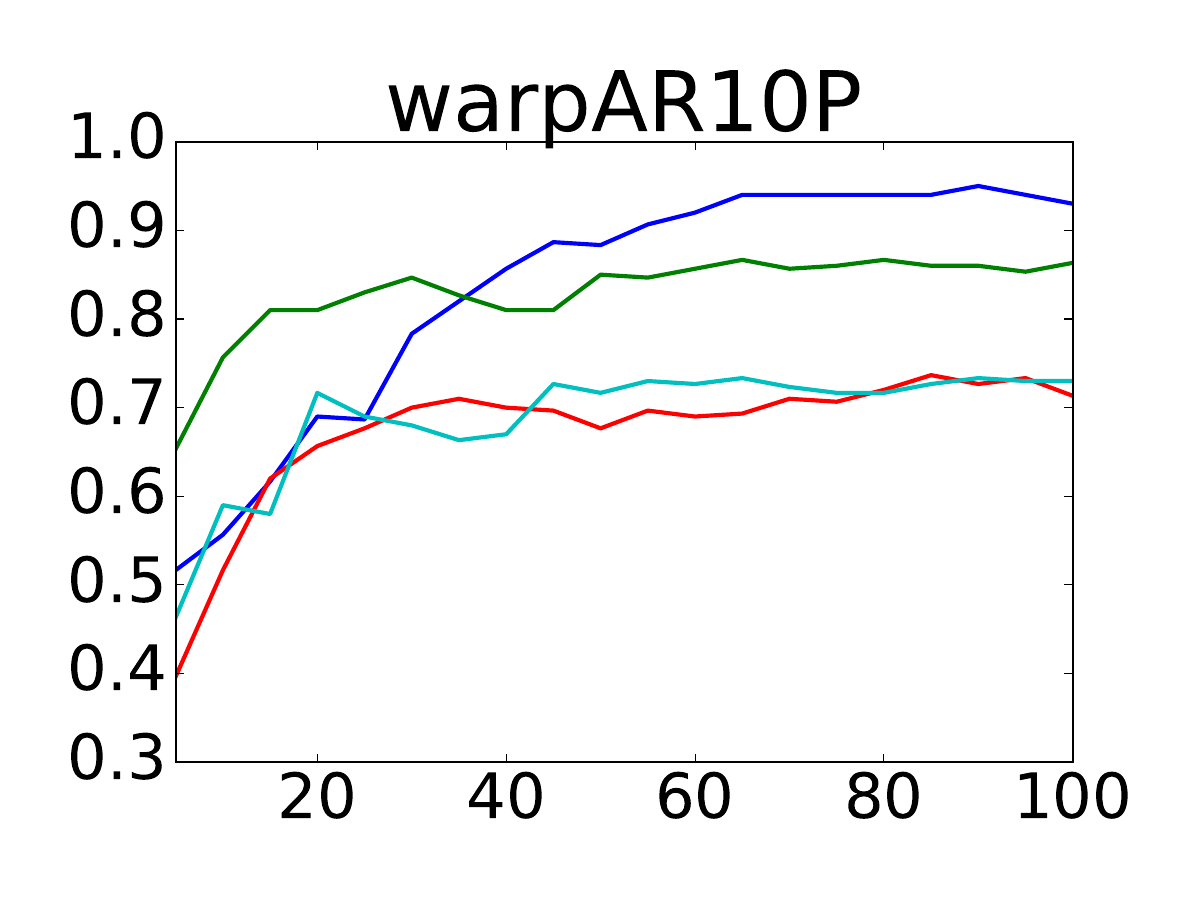}
\includegraphics[width=0.25\linewidth]{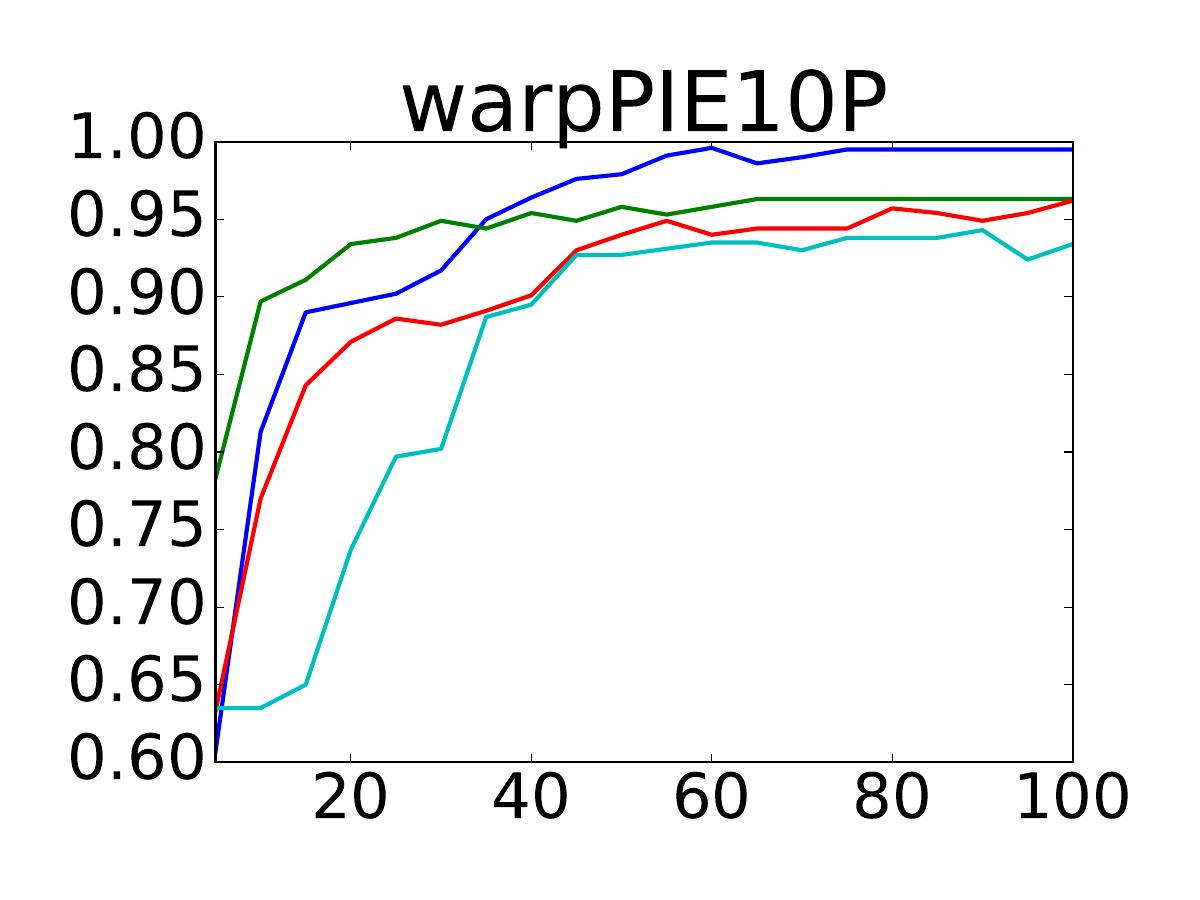}
\includegraphics[width=0.25\linewidth]{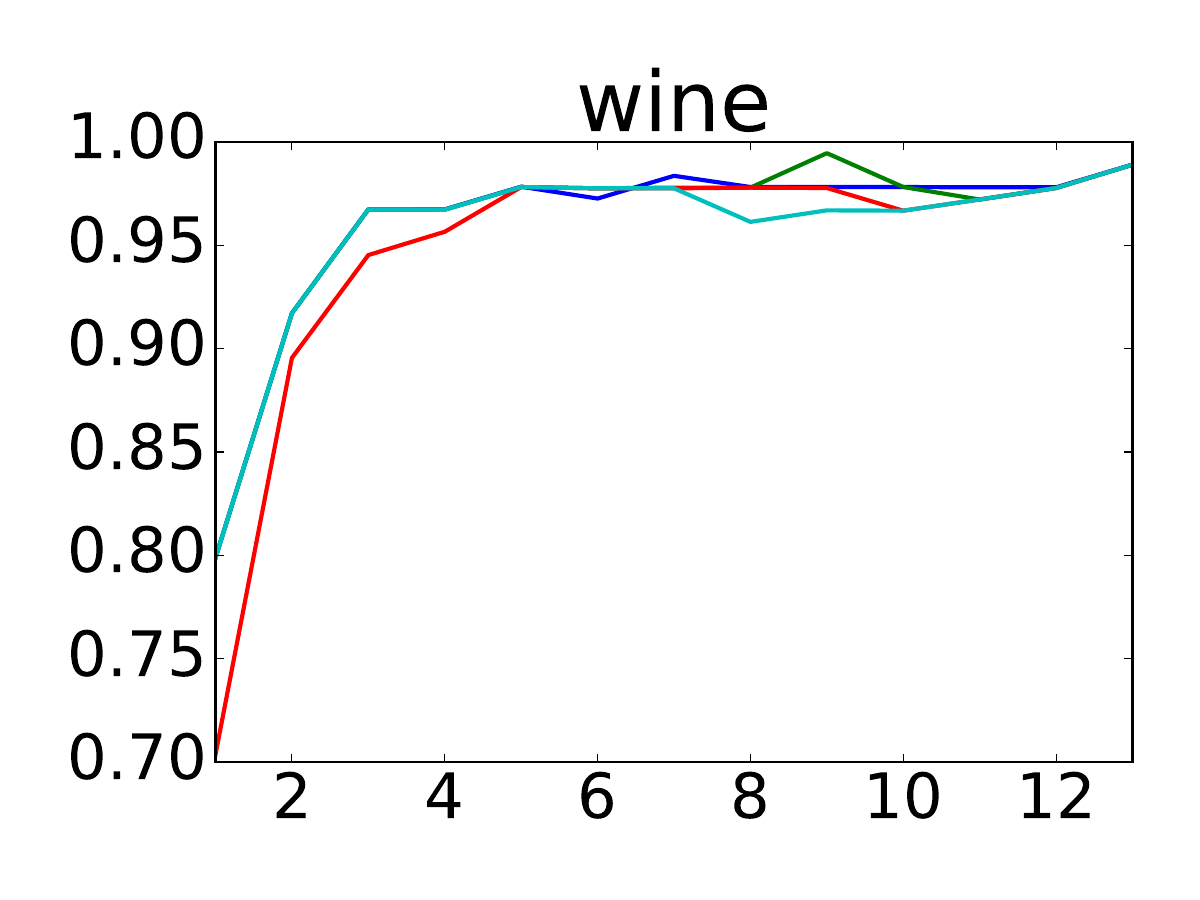}%
\includegraphics[width=0.25\linewidth]{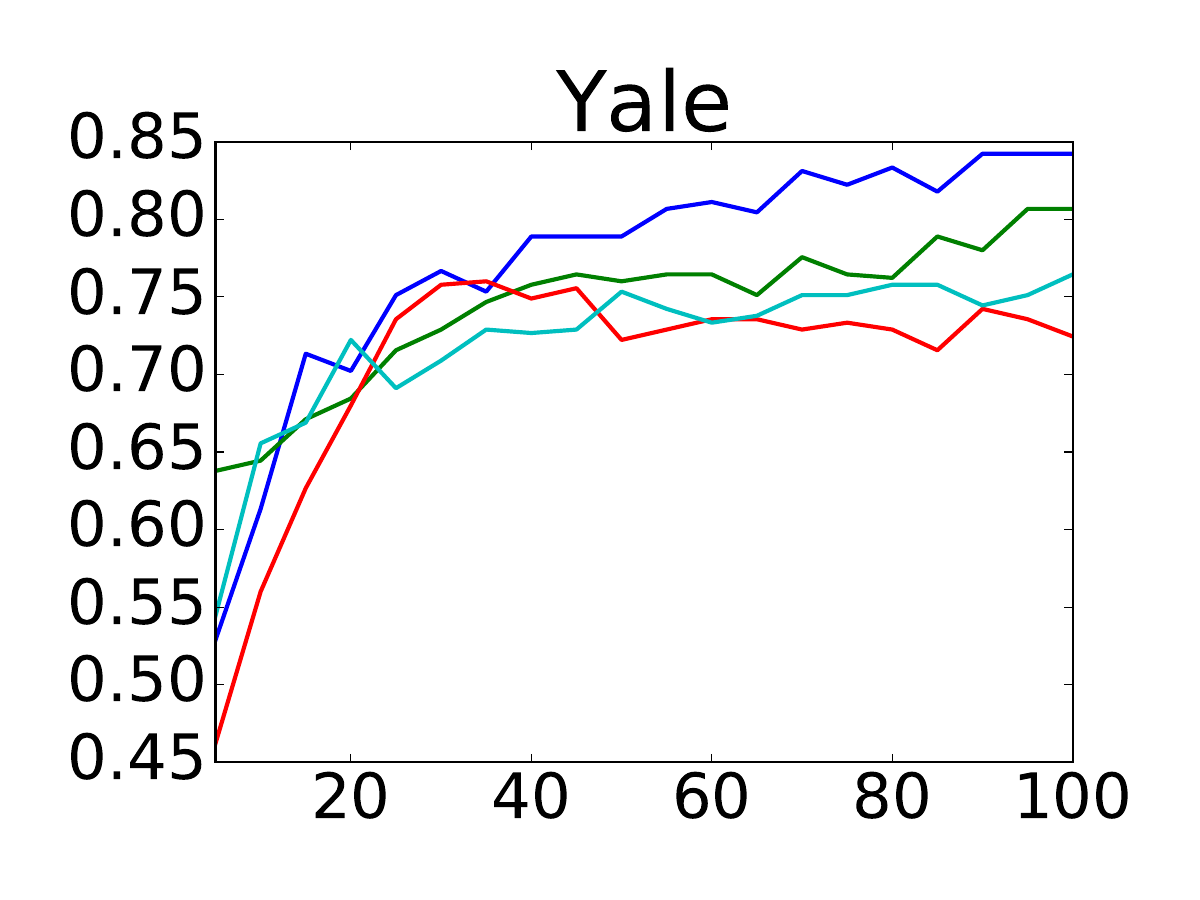} \\
\includegraphics[height=1.25em]{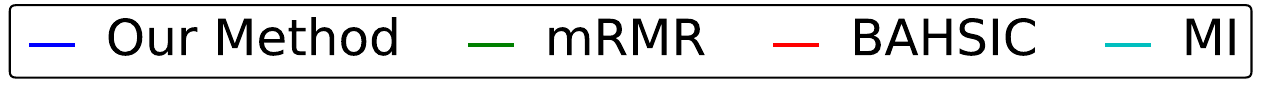}
\caption{The above plots show classification accuracy ($y$-axis)
  versus number of selected features ($x$-axis) for our real-world
  benchmark data sets. Higher accuracies are better.}
\label{fig:results-real-world}
\end{figure}

For every task, we run each algorithm being evaluated to obtain ranks
for all features. Performance is then measured by training a kernel
SVM on the top $m$ features and computing the resulting accuracy as
measured by 5-fold cross-validation. This is done for $m \in \{5, 10,
\dots, 100\}$ if the total number of features $d$ is larger than 100,
or $m \in \{1, 2, \dots, d\}$ otherwise. In all cases we fix the
regularization constant of the SVM to $C=1$ and use a Gaussian kernel
with $\sigma$ set as in the previous section over the selected
features. For our own algorithm, we fix $\varepsilon = 0.001$ across
all experiments and set the number of desired features to $m = 100$ if
$d > 100$ or $\lceil d/5 \rceil$ otherwise. Our results are shown in
Figure~\ref{fig:results-real-world}.

Compared with three other popular methods for nonlinear feature
selection, i.e.\ mRMR, BAHSIC, and MI, we find that our method is the
strongest performer in the large majority of cases, sometimes by a
substantial margin as in the case of \texttt{TOX-171}. While our
method is occasionally outperformed in the beginning when the number
of selected features is small, it either ties or overtakes the leading
method by the end in all but one instance. We remark that our method
consistently improves upon the performance of the related BAHSIC
method, suggesting that our objective based on conditional covariance
may be more powerful than one based on the Hilbert-Schmidt
independence criterion.

\begin{table}[t]
\centering

\tabcolsep=0.1cm
\resizebox{\linewidth}{!}{
\begin{tabular}{c|cccccccccccc}
& \texttt{ALLAML} & \texttt{CLL-SUB-111} & \texttt{glass} & \texttt{ORL} & \texttt{orlraws10P} & \texttt{pixraw10P} & \texttt{TOX-171} & \texttt{vowel} & \texttt{warpAR10P} & \texttt{warpPIE10P} & \texttt{wine} & \texttt{Yale} \\ \hline
Samples & 72 & 111 & 214 & 400 & 100 & 100 & 171 & 990 & 130 & 210 & 178 & 165 \\
Features & 7,129 & 11,340 & 10 & 1,024 & 10,304 & 10,000 & 5,784 & 10 & 2,400 & 2,420 & 13 & 1,024 \\
Classes & 2 & 3 & 6 & 40 & 10 & 10 & 4 & 11 & 10 & 10 & 3 & 15 \\
\end{tabular}
}

\vspace{1em}
\caption{Summary of the benchmark data sets we use for our empirical
  evaluation.}
\label{tab:real-data-summary}
\end{table}

\section{Conclusion}

In this paper, we proposed an approach to feature selection based on
minimizing the trace of the conditional covariance operator. The idea
is to select the features that maximally account for the dependence of
the response on the covariates.  We do so by relaxing from an
intractable discrete formulation of the problem to a continuous
approximation suitable for gradient-based optimization. We demonstrate
the effectiveness of our approach on multiple synthetic and real-world
experiments, finding that it often outperforms other state-of-the-art
approaches, including another competitive kernel feature selection
method based on the Hilbert-Schmidt independence criterion.

\bibliography{kfs}
\bibliographystyle{plainnat}
\clearpage

\appendix

\section{Appendix}
\label{appendix}

The appendix is devoted to proofs of various results from the main
text.


\subsection{Proof of Lemma \ref{lemma:characteristic}}

Suppose there exist two distributions $P,Q$ on $\R^m$ such that
$\E_{P(X)}[\ktil(y,X)]=\E_{Q(X)}[\ktil_1(y,X)]$ for any $y\in
\R^m$. Consider $\R^m$ as a subspace embedded in $\R^d$. The
probability distributions $P$ and $Q$ can be extended to $\R^d$ by by
setting the remaining components to zero. Then we also have
$\E_{P(X)}[\ktil_1(y,X)]=\E_{Q(X)}[\ktil_1(y,X)]$ for any
$y\in\R^d$. As $k_1$ is characteristic, $P=Q$.


\subsection{Proof of Theorem \ref{theorem:general} and Corollary \ref{cor:univariate}}
Theorem \ref{theorem:general} and Corollary \ref{cor:univariate} can
be proved simultaneously. The proof of Theorem \ref{theorem:general}
parallels the proof of the corresponding theorem in the setting of
dimension dimension reduction by \citet{fukumizu2009kernel}.

We can interpret $\tilde{\Hil}_1$ as a subset of $\Hil_1$, so Equation
\ref{prop:residual} implies $\Sigma_{YY|X_\T}\geq \Sigma_{YY|X}$. In
the univariate case, this is equivalent to saying
$\trace[\Sigma_{YY|X_\T}]\geq \trace[\Sigma_{YY|X}]$.

By the law of total variance, we have for any $g \in \Hil_2$,
\begin{align}\label{equation:total}
    \E_{X_\T}\var_{Y|X_\T}[g(Y)|X_\T]=\E_X[\var[g(Y)|X]]+
    \E_{X_\T}\var_{Y|X_\T}[\E_{Y|X}[g(Y)|X]].
\end{align}
By Lemma~\ref{lemma:characteristic}, the kernel $\ktil_1$ is
characteristic, so the conditional covariance operator characterizes
the conditional dependence, which reduces Equation
\ref{equation:total} to
\begin{align}
\langle g, (\Sigma_{YY|X_\T}-\Sigma_{YY|X})g\rangle =
\E_{X_\T}\var_{Y|X_\T}[\E_{Y|X}[g(Y)|X]] .
\end{align}  
Hence $\Sigma_{YY|X_\T}=\Sigma_{YY|X}$ if and only if given $X_\T$,
$\E_{Y|X}[g(Y)|X]$ is almost surely determined. Because $k_2$ is
characteristic, we have $Y\independent X|X_\T$. Suppose $Y$ is
univariate and $k_2$ is the linear kernel. Then both $\Sigma_{YY|X}$
and $\Sigma_{YY|X_\T}$ can be equivalently interpreted as linear
functions that map real numbers to real numbers. When $g=\text{Id}_Y$,
the identity function on $Y$, we have
\begin{align}
  \langle \text{Id}_Y,
  (\Sigma_{YY|X_\T}-\Sigma_{YY|X})\text{Id}_Y\rangle =
  \E_{X_\T}\var_{Y|X_\T}[\E_{Y|X}[Y|X]] = 0.
\end{align}
This implies $Y\independent X|X_\T$ directly.

\subsection{Proof of Theorem~\ref{theorem:consistency}}

We provide a simpler proof than the one for Theorem 6 in the
paper~\cite{fukumizu2009kernel}, where the consistency result for
dimension reduction was established.

For any subset of features $\T$, we have
\begin{align*}
& |\trace[\hat\Sigma_{YY|X_\T}]-\trace[\Sigma_{YY|X_\T}] | \\
& \leq |
  \trace[\Sigma_{YY|X_\T}]-\trace[\Sigma_{YY}-\Sigma_{YX_\T}(\Sigma_{X_\T
      X_\T} + \varepsilon_n I)^{-1} \Sigma_{X_\T Y}] | + \\
& + |\trace[\Sigma_{YY}-\Sigma_{YX_\T}(\Sigma_{X_\T X_\T} +
    \varepsilon_n I)^{-1} \Sigma_{X_\T Y}] - \trace [ \hat
    \Sigma_{YY|X_\T}] |,
\end{align*}
where the second term converges to zero by the law of large numbers,
whereas \citet{fukumizu2009kernel} proved that the second term can be
upper bounded as
\begin{align*}
 \frac{1}{\varepsilon_n}\{ &
 (\|\hat\Sigma_{YX_\T^{(n)}}\|_{\text{\text{HS}}} +
 \|\Sigma_{YX_\T}\|_{HS}) \|\hat\Sigma_{YX_\T} -
 \Sigma_{YX_\T}\|_{\text{HS}} \\
& +\|\Sigma_{YY}\|_{\trace}\|\hat\Sigma_{X_\T X_\T}^{(n)} -
 \Sigma_{X_\T X_\T}\|_{\text{HS}} \\
& + |\trace[ \hat\Sigma_{YY}-\Sigma_{YY}] | \},
\end{align*}
where $\|\cdot\|_{\text{HS}}$ is the HSIC norm of an operator. By the
Central Limit Theorem, both of the terms
\begin{align*}
\|\hat\Sigma_{YX_\T}-\Sigma_{YX_\T}\|_{\text{HS}},\|\hat\Sigma_{X_\T
  X_\T}^{(n)}-\Sigma_{X_\T X_\T}\|_{\text{HS}} \quad \mbox{and} \quad
|\trace[\hat\Sigma_{YY}-\Sigma_{YY}]|
\end{align*}
are guaranteed to be of order $\order_p(n^{-1/2})$. Hence, the second
term also converges to $0$. This establishes the convergence of
$\trace[\hat\Sigma_{YY|X_\T}]$ towards $\trace[\Sigma_{YY|X_\T}]$,
which yields the claim~\eqref{consistency} by standard
$\varepsilon$--$\delta$ arguments.

\end{document}